\documentclass[runningheads]{llncs}
\usepackage{graphicx}
\usepackage{multirow}
\usepackage{xcolor}
\usepackage{amsmath,amssymb}
\begin{document}
\title{Multi-Task Learning for User Engagement and Adoption in Live Video Streaming Events}
\titlerunning{MERLIN for User Engagement and Adoption in Live Video Streaming}
%
\author{Stefanos Antaris \inst{1,2} \and Dimitrios Rafailidis \inst{3} \and Romina Arriaza \inst{2}}
%
%
\institute{KTH Royal Institute of Technology, Sweden  \and
Hive Streaming AB, Sweden \\ 
\email{antaris@kth.se} \\
 \email{romina.arriaza@hivestreaming.com} \and
University of Thessaly, Greece \\
\email{draf@uth.gr}}
\maketitle              
\begin{abstract}

Nowadays, live video streaming events have become a mainstay in viewer's communication in large international enterprises. Provided that viewers are distributed worldwide, the main challenge resides on how to schedule the optimal event's time so as to improve both the viewer's engagement and adoption. In this paper we present a multi-task deep reinforcement learning model to select the time of a live video streaming event, aiming to optimize the viewer's engagement and adoption at the same time. We consider the engagement and adoption of the viewers as independent tasks and formulate a unified loss function to learn a common policy. In addition, we account for the fact that each task might have different contribution to the training strategy of the agent. Therefore, to determine the contribution of each task to the agent's training, we design a Transformer's architecture for the state-action transitions of each task. We evaluate our proposed model on four real-world datasets, generated by the live video streaming events of four large enterprises spanning from January 2019 until March 2021. Our experiments demonstrate the effectiveness of the proposed model when compared with several state-of-the-art strategies. For reproduction purposes, our evaluation datasets and implementation are publicly available at \url{https://github.com/stefanosantaris/merlin}.



\end{abstract}
\section{Introduction} \label{sec:intro}


Over the last years, video streaming technologies have been widely exploited by large international enterprises as the main internal communication medium \cite{ibmvideo}. The enterprises schedule several live video streaming events to communicate with thousands of their employees, who are spread around the world. To ensure that every employee/viewer attends the event without experiencing poor network performance, the enterprises exploit distributed live video streaming solutions. Such solutions account for each office's internal bandwidth to overcome network congestion and distribute the streaming video to viewers \cite{antaris2020vstreamdrls}. Although distributed solutions ensure that every viewer can attend the event, an erroneously scheduled time of an event negatively affects the viewer's engagement, that is the percentage of the event's duration that a viewer attends \cite{hbr}. In practice, the viewers partially attend  the entire duration of an event, when an event is erroneously scheduled on a non-preferred time e.g., day and hour, resulting in a low viewer's engagement. Moreover, the erroneously scheduled time impacts the number of enterprise's events that each viewer participates, reflecting on the viewer's adoption. In particular, the viewers with several time zones have low adoption, when organizing the events and ignoring the viewer's availability. Instead of manually organizing the events, it is important for the enterprises to develop a mechanism to learn how to schedule an event on the day and hour that optimizes both the viewer's engagement and adoption.

To organize an event, enterprises interact with a centralized agent that is located in a company  offering the live video streaming solution. However, current streaming solutions do not account for the optimal selection of the time of the next event. To overcome the shortcomings of current live video streaming solutions, in this study we follow a reinforcement learning strategy and design an agent that receives the viewer's engagement and adoption as two different reward signals for the selection of the event's time. Reinforcement learning has been proven an efficient means for optimizing a reward signal in various domains such as robotics \cite{polydoros2017survey,Zhu2020The}, games \cite{silver2017mastering,Ye2020aaai}, recommendation systems \cite{Liu2020wsdm,Xin2020sqn}, and so on. However, such approaches train an agent on a single task, where the learned policy maximizes a single cumulative reward. Nonetheless, the goal of the agent in our case of the event's time selection problem is to optimize both the viewer's engagement and adoption rewards.  Recently, multi-task reinforcement learning approaches have been proposed to generate a single agent that learns a policy which optimizes multiple tasks, with each task corresponding to a different reward signal \cite{lasse2018impala,Hessel2019Popart,Teh2017distral}. State-of-the-art approaches train an agent by sharing knowledge among similar tasks \cite{Varghese2020survey}. For example, the attentive multi-task deep reinforcement learning (AMT) model~\cite{Timo2020AMT} exploits a soft-attention mechanism to train a single agent on tasks that follow different distributions in the reward signal. However, AMT transfers knowledge among similar tasks, while isolating dissimilar tasks during the agent's training. This means that AMT achieves sub-optimal performance when tasks have completely different characteristics, as it happens in the case of live video streaming events. For instance, as we will demonstrate in Section \ref{sec:data} the viewers have a low engagement behavior over time, whereas the viewer's adoption increases among consecutive events. 

In addition, to efficiently select the event's time, the agent has to capture the evolution of the viewer's engagement and adoption. Towards this aim, the Transformer's architecture has been emerged as a state-of-the-art learning model across a wide variety of evolving tasks \cite{vaswani2017attention}. For example, in \cite{parisotto20transformer} the Transformer's architecture has been exploited in a reinforcement learning strategy to provide memory to the agent by preserving the sequence of the past observations. However, baseline approaches based on the Transformer's architecture have not been studied for multi-task reinforcement learning problems. 

To address the shortcomings of state-of-the-art strategies, in this study we propose a \textbf{M}ulti-task l\textbf{E}a\textbf{R}ning model for user engagement and adoption in \textbf{L}ive v\textbf{I}deo streami\textbf{N}g events (MERLIN), making the following contributions:
\begin{itemize}
    \item We formulate the viewer's engagement and adoption tasks as different Markov Decision Processes (MDPs) and propose a multi-task reinforcement learning strategy to train an agent that selects the optimal time, that is day and hour of the enterprise's next event aiming to maximize both tasks.
    \item We design a Transformer's architecture to weigh the importance of each task during the training of the agent, that is to determine the contribution of each task to the learning strategy of the agent's policy. 
    \item We transfer knowledge among tasks through a joint loss function in a multi-task learner component and compute a common policy that optimizes both the viewer's engagement and adoption in a live video streaming event.
\end{itemize}
Our experimental evaluation on four real-world datasets with live video streaming events show the superiority of the proposed MERLIN model over baseline multi-task reinforcement learning strategies. The remainder of this paper is organized as follows, in Section \ref{sec:data} we present the main characteristics of the live video streaming events as well as the evolution of the viewer's engagement and adoption. In Section \ref{sec:proposed} we formally define the multi-task problem of scheduling live video streaming events, and  detail the proposed MERLIN model. Then, in Section \ref{sec:exp} we present the experimental evaluation of our model against baseline strategies, and conclude the study in Section \ref{sec:conclusion}.

\section{Live Video Streaming Events} \label{sec:data}

\begin{table}[]
\caption{Statistics of the datasets with all the live video streaming events that took place in four international enterprises from January 2019 until March 2021.}
\resizebox{\textwidth}{!}{
    \centering
    
    \begin{tabular}{c|c|c|c|c}
    \hline
         &  \textbf{Enterprise 1 (E1)} & \textbf{Enterprise 2 (E2)} & \textbf{Enterprise 3 (E3)} & \textbf{Enterprise 4 (E4)} \\ \hline
        \textbf{\#Events} & $833$ & $1,303$ & $3,025$ & $7,249$ \\ \hline
        \textbf{\#Viewers} & $98,296$ & $59,090$ & $194,026$ & $508,654$ \\ \hline
        \textbf{\#Time zones} & $63$ & $97$ & $167$ & $150$ \\ \hline
        \textbf{Avg. Engagement} ($u_t$) & $0.455$ & $0.422$ & $0.383$ & $0.409$ \\ \hline
        \textbf{Avg. Adoption} ($v_t$) & $1.275$ & $6.905$ & $8.528$ & $6.375$ \\ \hline
    \end{tabular}
}
    \label{tab:stats}
\end{table}

\begin{figure*}
    \centering
    \includegraphics[scale=0.118]{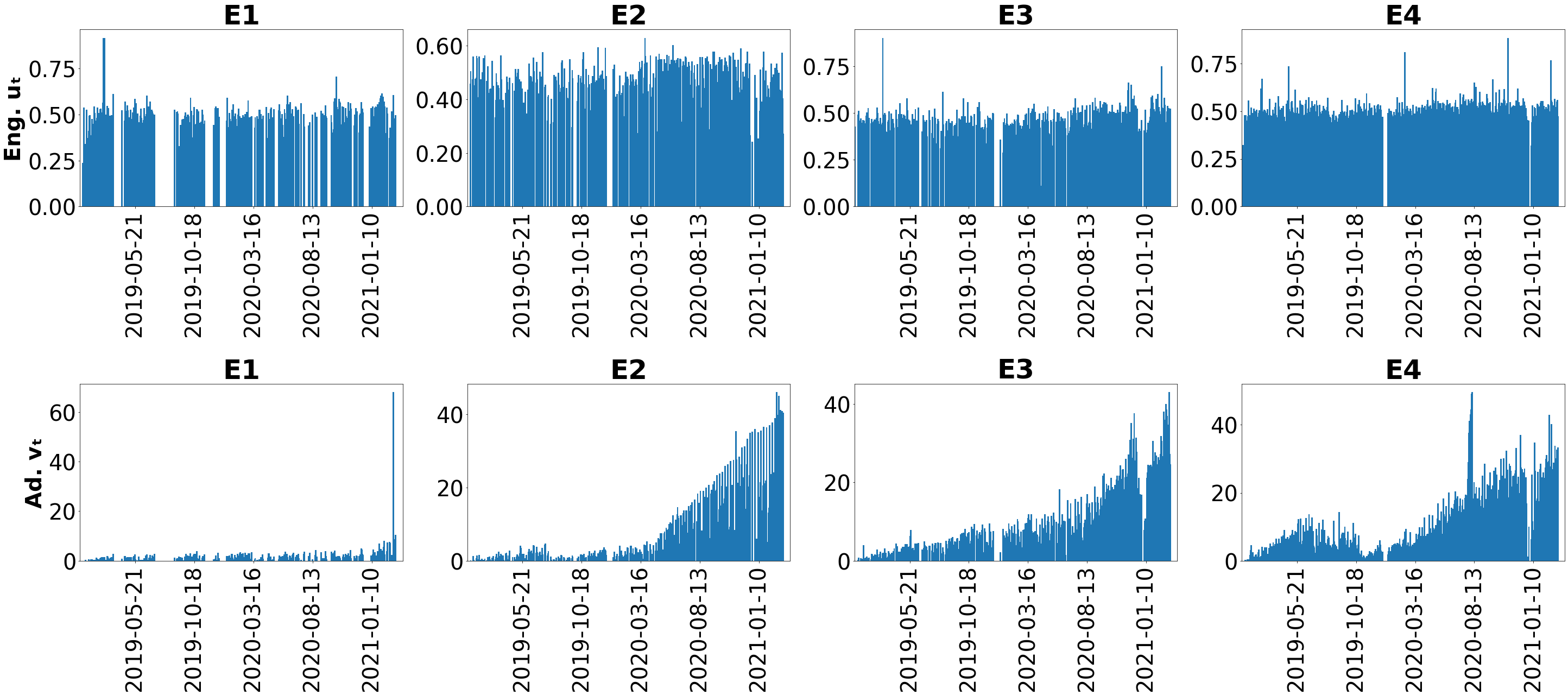}
    \caption{Evolution of viewer's engagement $u_t$ and adoption $v_t$ in the events.}
    \label{fig:engagement}
\end{figure*}

We collected four real-world datasets with all the events that occurred in four large enterprises worldwide from January 2019 until March 2021. The video streaming solution of the events was supported by our company. We monitored a set $\mathcal{E}$ of live video streaming events, where for each event $e_t \in \mathcal{E}$ on date $t$ the viewers reported to a backend server of our company the timezones, as well as their joining and leaving times during the event. The datasets were anonymized and made publicly available. In Table \ref{tab:stats}, we summarize the statistics of the four evaluation datasets. Each enterprise has a different number of viewers, located in several countries around the world with different time zones. We observe that the viewers in Enterprise 1 are distributed to less time zones than the other enterprises, whereas Enterprise 4 hosts the largest number of live video streaming events with approximately 0.5M viewers in total. In Figure \ref{fig:engagement}, we present the average viewer's engagement to the live video streaming events throughout the time span. We define the average engagement $u_t$ of the viewers that participated in the event $e_t \in \mathcal{E}$ on the date $t$ as follows:
\begin{equation}
u_t = \frac{1}{n} \sum_{i=1}^{n} \frac{k_i}{m}
\label{eq:engagement}    
\end{equation}
where $n$ is the number of viewers that participated in the event $e_t$, $k_i$ is each viewer's attendance time and $m$ is the duration of the event. In all enterprises the viewers have low engagement, that is in all enterprises the viewers attended less than the half duration of each live video streaming event with average viewer's engagement $u_t<$0.5  (Table~\ref{tab:stats}). In addition, the average viewer's adoption expresses how many events the viewers attended until a date $t$, where large adoption scores indicate that viewers were willing to participate in the enterprise's previous events. We formally define the average adoption $v_t$ as follows:
\begin{equation}
    v_t = \frac{\sum_{i=1}^{n} c_i}{n}
    \label{eq:adoption}
\end{equation}
where $c_i$ is the number of events that each viewer $i$ attended prior to the event $e_t$.  We observe that the viewers in Enterprise 1 adopted less events than the other enterprises with average adoption $v_t$=1.275. On one of the last dates Enterprise 1 organized an all-hands event where all the viewers were invited, which explains the pick of the adoption score for Enterprise 1 in Figure \ref{fig:engagement}. The adoption scores for Enterprises 2, 3 and 4 increase over time in the last year, as enterprises started to organize more events than the previous years for viewers who most of them  worked from home due to the COVID'19 pandemic.

\section{Proposed Model} \label{sec:proposed}

An enterprise organizes $T = |\mathcal{E}|$ events, where each event on a date/step $t$ is defined as $e_t = (h, n, m, u_t, v_t, \mathbf{z})$, with $h$ being a timestamp that corresponds to the event's day and hour. Notice that a date/step $t$ has 24 different timestamps $h$ and an event $e_t$ has a duration of $m$ minutes with $n$ viewers. The viewers attend the event with different time zones which is represented as an one-hot vector $\mathbf{z} \in \mathbb{R}^{d_z}$, where $d_z$ is the number of different time zones of the viewers. The goal of the enterprise is to organize each event $e_t \in \mathcal{E}$ on the timestamp $h$, to maximize the average engagement $u_t$ and adoption $v_t$ of the viewers. We formulate the scheduling of the next event as a Markov Decision Process (MDP), where the agent interacts with the environment/enterprise by selecting the timestamp $h$ of the next event $e_{t+1}$ and maximizing the cumulative rewards. In particular, we define the MDP of the live video streaming event as follows \cite{sutton2018reinforcement}: 

\begin{definition} \textbf{Live Video Streaming Event MDP.}
At each step $t = 1, \ldots, T$, the agent interacts with the environment and selects an action $\mathbf{a}_t \in \mathcal{A}$. An action $\mathbf{a}_t$ corresponds to the selection of the timestamp $h$ of the next event $e_{t+1}$ based on the state $\mathbf{s}_t \in \mathcal{S}$ of the enterprise. We define the state $\mathbf{s}_t$ of the enterprise as a sequence of the $l$ previous events $\mathbf{s}_t = \{e_{t-l}, \ldots, e_t\}$\footnote{We consider only the $l$ previous events to capture the most recent viewers behavior. As we will demonstrate in Section \ref{sec:exp}, considering large values of $l$ does not necessarily improve the model's performance.}. The agent receives a reward $r(\mathbf{s}_t, \mathbf{a}_t,\mathbf{s}_{t+1}) \in \mathcal{R}$ for selecting the action $\mathbf{a}_t \in \mathcal{A}$ in state $\mathbf{s}_t \in \mathcal{S}$, considering the enterprise transitions to state $\mathbf{s}_{t+1}$ with a probability $p(\mathbf{s}_{t+1}|\mathbf{s}_t,\mathbf{a}_t) \in \mathcal{P}$. The goal of the agent is to find the optimal policy $\pi_{\theta}: \mathcal{S} \times \mathcal{A} \rightarrow \mathcal{R}$, where $\theta$ is the set of policy parameters, assigning a probability $\pi_{\theta}(\mathbf{a}_t|\mathbf{s}_t)$ of selecting an action $\mathbf{a}_t \in \mathcal{A}$ provided a state $\mathbf{s}_t \in \mathcal{S}$. Having computed the policy $\pi_{\theta}$, the agent maximizes the expectation of the discounted cumulative reward $\max \mathbb{E}[\sum_{t=0}^{T}\gamma^{t}r(\mathbf{s}_t,\mathbf{a}_t,\mathbf{s}_{t+1})| \pi_{\theta}]$, with $\gamma \in [0,1]$ being the discount factor.
\end{definition}

In our model, we focus on training a common agent that optimizes both the viewer's engagement $u_t$ and adoption $v_t$. As mentioned in Section \ref{sec:data}, the viewer's engagement and adoption behavior vary over time. Therefore, we first consider the viewer's engagement and adoption as independent tasks, and then train a common agent to optimize the cumulative rewards of both tasks at the same time. We define the multi-task Reinforcement Learning (RL) problem in live video streaming events as follows \cite{Timo2020AMT,Calandriello2014Nips,lasse2018impala,Hessel2019Popart}:

\begin{definition} \textbf{Multi-Task RL in Live Video Streaming}
In the multi-task RL problem for live video streaming events, we consider a set of tasks $\mathcal{T}$, that is the engagement and adoption tasks with $|\mathcal{T}|=2$. We formulate each task $\tau \in \mathcal{T}$ as a different MDP, where the tasks have the same state $\mathcal{S}$ and action space $\mathcal{A}$ with a different set of rewards $\mathcal{R}$. For the engagement task we compute reward $r(\mathbf{s}_t, \mathbf{a}_t)$ as the average engagement $u_t$ in Equation \ref{eq:engagement}, and for the adoption task the reward corresponds to the average adoption $v_t$ in Equation \ref{eq:adoption} at the $t$-th step. The goal of the agent is to learn a common policy $\pi_{\theta}$ that solves each task $\tau \in \mathcal{T}$, by maximizing the expected return $\max \mathbb{E}_{\tau \sim \mathcal{T}}[[\sum_{t=0}^{T}\gamma^{t}r(\mathbf{s}^{\tau}_t,\mathbf{a}^{\tau}_t,\mathbf{s}^{\tau}_{t+1})| \pi_{\theta}]]$ for both tasks. $\mathbf{s}^{\tau}_t$ is the state of the agent and $\mathbf{a}^{\tau}_t$ is the action taken by the agent for the task $\tau$ at the $t$-th step.
\end{definition}

\subsection{MERLIN's Architecture}
\label{sec:arc}

\begin{figure*}[ht]
    \centering
    \includegraphics[scale=0.3]{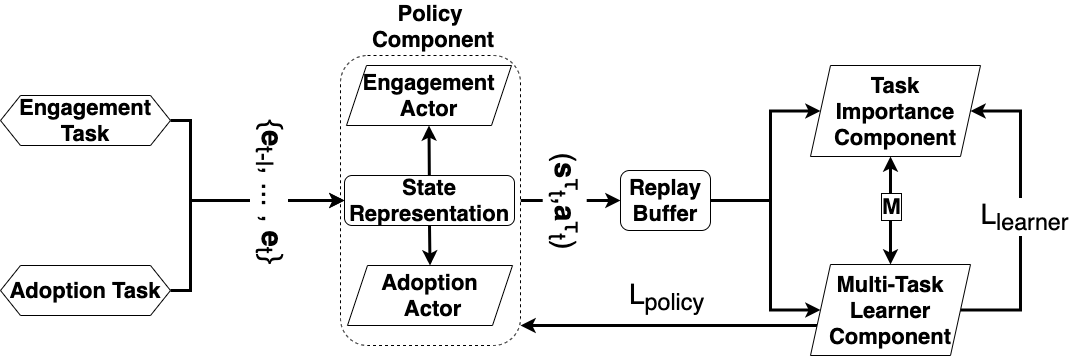}
    \caption{The architecture of the proposed MERLIN model for the viewer's engagement and adoption tasks. MERLIN consists of: (i) the policy, (ii) task importance and (iii) multi-task learner components.}
    \label{fig:arch}
\end{figure*}

As illustrated in Figure \ref{fig:arch}, the proposed MERLIN model consists of three main components: the policy, task importance and multi-task learner components. The goal of MERLIN is to compute a common policy $\pi_{\theta}$ that maximizes the future rewards for the viewer's engagement and adoption tasks $\tau \in \mathcal{T}$.

\noindent \textbf{- Policy Component}. The role of the policy component is to compute the action $\mathbf{a}^{\tau}_t$ of both tasks. During training, the agent interacts with two environments in the enterprise, that is the different two tasks $\tau \in \mathcal{T}$. The input of the policy component is the $l$ previous events $\{e^{\tau}_{t-l}, \ldots, e^{\tau}_t\}$ of each task. We implement a shared state representation module to compute the state $\mathbf{s}^{\tau}_t$ of task $\tau$. In our architecture, we design the respective two actors to generate the actions $\mathbf{a}^{\tau}_t$ for the engagement and adoption tasks \cite{lasse2018impala}. Then, the generated state-action transitions by both actors are stored in the replay buffer with size $l_b$ to train the common agent.

\noindent \textbf{- Task Importance Component}. The task importance component determines the contribution of each task to the learning process of the agent. Notice that state-of-the-art RL strategies are designed to learn a policy of a single agent that optimizes similar tasks, ignoring the information of each task's state-action transition \cite{Varghese2020survey}. Instead, in the proposed MERLIN model to account for the impact of each state-action transition on the policy $\pi_{\theta}$, we consider the encoder model of the Transformer's architecture for the state-action transition sequences. In doing so, we capture the information of the state-action transitions of both the engagement and adoption actors over time \cite{loynd2020working,parisotto20transformer}. In addition, the task importance component computes a weight matrix $\mathbf{M} \in \mathbb{R}^{l_b \times |\mathcal{T}|}$ which reflects on the contribution of each actor to the learning process of the policy $\pi_{\theta}$.

\noindent \textbf{- Multi-Task Learner Component}. The role of the multi-task learner component is to optimize the policy $\pi_{\theta}$ based on the $l_b$ state-action transitions stored in the replay buffer. Provided the stored state-action transitions in the replay buffer and the weight matrix $\mathbf{M}$ of the task importance component, the multi-task learner updates the policy parameters through a joint loss function $\mathcal{L}_{policy}$ and the parameters of the task importance component via the $\mathcal{L}_{learner}$ function, following the temporal-difference learning strategy \cite{sutton2018reinforcement}. In particular, matrix $\mathbf{M}$ first weighs the state-action transitions in the replay buffer, and then the multi-task learner optimizes the joint loss function $\mathcal{L}_{policy}$ to compute the parameters of the policy component. In addition, the multi-task learner learns its parameters via the joint loss function  $\mathcal{L}_{learner}$, and updates the parameters of the task importance component accordingly.

\subsection{Policy Component}

At each step $t = 1, \ldots, T$, the policy component takes as an input the $l$ previous events $\{e^{\tau}_{t-l},\ldots, e^{\tau}_t\}$ of each task $\tau \in \mathcal{T}$. The goal of the policy component is to learn a policy $\pi_{\theta}$ that solves each task $\tau$. Provided that the engagement and adoption tasks have the same state space $\mathcal{S}$ and action space $\mathcal{A}$, the policy component consists of a shared state representation module and two actors, that is the engagement and adoption actors.

\noindent \textbf{- State Representation Module}. The state representation module takes as an input the $l$ previous events $\{e^{\tau}_{t-l},\ldots, e^{\tau}_t\}$, and generates the state $\mathbf{s}^{\tau}_t$ of each task $\tau$ at the $t$-th step. We represent each event $e^{\tau}_t$ as a $d_x$-dimensional vector $\mathbf{x}^{\tau}_t \in \mathbb{R}^k$ concatenating the event's features $\mathbf{x}^{\tau}_t = Concat(h, n, m, g, o, \mathbf{z})$. Given the representations $\{\mathbf{x}^{\tau}_{t-l}, \ldots, \mathbf{x}^{\tau}_t\}$ of the $l$ previous events, we compute the $d_s$-dimensional state representation vector $\mathbf{s}^{\tau}_t \in \mathbb{R}^{d_s}$ as follows: \cite{Zhu2017lstm,zou2019reinforcement}:

\begin{equation}
    \mathbf{s}^{\tau}_t = \xi_{w}(\mathbf{x}^{\tau}_t, \Delta(t)) = \textrm{Time-LSTM}(\mathbf{x}^{\tau}_t, \Delta(t))
    \label{eq:xiw}
\end{equation}
where $w$ are the trainable parameters of the Time-LSTM function $\xi(\cdot)$~\cite{Zhu2017lstm}. Notice that Time-LSTM models the time difference $\Delta(t)$ of the event $e^{\tau}_t$ and the previous event $e^{\tau}_{t-1}$ as follows:

\begin{equation}
\begin{array}{l}
    \mathbf{g}_t = \sigma \bigg(\mathbf{x}^{\tau}_t \mathbf{W}_{xg} + \sigma (\Delta(t) \mathbf{W}_{g} + b_{g}) \bigg) \\
    \mathbf{q}_t = \mathbf{f}_t \odot \mathbf{q}_{t-1} + \mathbf{i}_t \odot \mathbf{g}_t \odot \sigma \bigg( \mathbf{x}^{\tau}_t \mathbf{W}_{xq} + \mathbf{s}_{t-1} \mathbf{W}_{sq} + b_q \bigg) \\
    \mathbf{o}_t = \sigma(\mathbf{x}^{\tau}_t \mathbf{W}_{xo} + \Delta(t) \mathbf{W}_{o} + \mathbf{s}^{\tau}_{t-1} \mathbf{W}_{so} + \mathbf{q}_t \odot \mathbf{W}_{qo} + b_o) \\
    \mathbf{s}^{\tau}_t = \mathbf{o}_t \odot \sigma(\mathbf{q}_t)
\end{array}
\label{eq:time_lstm}
\end{equation}
where $\mathbf{g}_t$ is the time dependent gate influencing the memory cell and the output gate $\mathbf{o}_t$, $\mathbf{q}_t$ is the memory cell of LSTM, and $\mathbf{f}_t$ and $\mathbf{i}_t$ are the forget and input gates, respectively \cite{graves2013generating,zou2019reinforcement}. The symbol $\odot$ represents the Hadamard element-wise product and $\sigma (\cdot)$ is the sigmoid function. The different weight matrices $\mathbf{W}_{*}$ in Equation 4 transform the event embedding $\mathbf{x}^{\tau}_t$ and the time difference $\Delta(t)$ to the $d_s$-dimensional latent space, and $b_{*}$ are the respective bias terms. Notice that the time difference $\Delta(t)$ is important to capture the similarity among consecutive events in the state $\mathbf{s}^{\tau}_t$. Provided that the engagement and adoption of the viewers vary over time, our goal is to capture the most recent viewer's behaviour in the state space $\mathbf{s}^{\tau}_t$. Therefore, the Time-LSTM in Equation \ref{eq:time_lstm} tends to forget events with high time difference, and focuses on the recent events.

\noindent \textbf{- Engagement and Adoption Actors}. The engagement and adoption actors take as input the state $\mathbf{s}^{\tau}_t$ of each task $\tau \in \mathcal{T}$. The state representation $\mathbf{s}^{\tau}_t$ captures the evolution of the enterprise over time. Given the state $\mathbf{s}^{\tau}_t$ and a policy $\pi_{\theta}$, each actor computes a $d_a$-dimensional action vector $\mathbf{a}^{\tau}_t \in \mathbb{R}^{d_a}$, where $d_a$ is the number of all the possible timestamps. Each dimension of the action vector $\mathbf{a}^{\tau}_t$ corresponds to the probability of selecting the timestamp $h$ for the next event $e_{t+1}$. We implement a two-layer perceptron (MLP) to transform the state vector $\mathbf{s}^{\tau}_t \in \mathbb{R}^{b}$ to the action vector $\mathbf{a}^{\tau}_t \in \mathbb{R}^{u}$ as follows:

\begin{equation}
    \mathbf{a}^{\tau}_t = \pi_{\theta}(\mathbf{s}^{\tau}_t) = MLP(\mathbf{s}^{\tau}_t)
    \label{eq:action}
\end{equation}
where $\theta$ are the trainable parameters of the MLP, that is the policy parameters of the agent. Given the action vector $\mathbf{a}^{\tau}_t$ of each actor, we normalize the action vector $\mathbf{a}^{\tau}_t$ based on the softmax function and select the action with the highest value using the $\epsilon$-greedy exploration technique \cite{sutton2018reinforcement}. The generated state-action transitions are stored in the replay buffer to learn the optimal policy $\pi_{\theta}$ based on the past experiences of each task.

\subsection{Task Importance Component}

The goal of the task importance component is to determine the contribution of each task to the learning strategy of the policy $\pi_{\theta}$. The input of the task importance component is the set of state-action transitions stored in the replay buffer by the engagement and adoption actors. At each step $t = 1, \ldots, T$, the engagement and adoption actors store in the replay buffer the respective  state-action transition $(\mathbf{s}^{\tau}_t, \mathbf{a}^{\tau}_t)$ of the task $\tau \in \mathcal{T}$. Having stored the $l_b$ state-action transitions of each task $\tau$ in the replay buffer, the task importance component computes the similarity among the tasks. As the replay buffer contains a sequence of state-action transitions, we employ the encoder of the Transformer's model to capture the information of the $l_b$ states to $d_y$-dimensional vectors $\mathbf{Y}^{\tau} \in \mathbb{R}^{l_b \times d_y}$ \cite{vaswani2017attention}. To overcome any stability problems that might occur at the early stages of the training, we implement the Gated Transformer(-XL) (GTrXL) model of the Transformer's architecture as follows \cite{parisotto20transformer}:

\begin{equation}
    \mathbf{Y}^{\tau} = \psi_{\eta}(\{\mathbf{s}^{\tau}_{t-l_b}, \ldots, \mathbf{s}^{\tau}_t\}) = GTrXL(\{\mathbf{s}^{\tau}_{t-l_b}, \ldots, \mathbf{s}^{\tau}_t\})
\end{equation}
where $\{\mathbf{s}^{\tau}_{t-l_b}, \ldots, \mathbf{s}^{\tau}_t\}$ is the states sequence of the task $\tau$ stored in the replay buffer. Parameters $\eta$ denote the trainable weights of the GTRrXL function $\psi(\cdot)$~\cite{parisotto20transformer}.

By computing the $d_y$-dimensional vectors, that is the rows of matrix $\mathbf{Y}^{\tau}$ of each task $\tau$, we deduce the importance of each state $\mathbf{s}^{\tau}_t$ in the actions selected by the actor over time for task $\tau$. Therefore, we can compute a weight matrix $\mathbf{M} \in \mathbb{R}^{l_b \times |\mathcal{T}|}$ of each state $\mathbf{s}^{\tau}_t$ during the training of the agent's policy $\pi_{\theta}$. To calculate the weight matrix $\mathbf{M}$, we employ a two-layer MLP with softmax activation:

\begin{equation}
    \mathbf{M} = \lambda_{\omega}(\mathbf{Y}^{\tau}) = softmax\bigg(MLP(\mathbf{Y}^{\tau})\bigg)
\end{equation}
where $\omega$ are the parameters of the MLP transformation function $\lambda(\cdot)$. Intuitively, we give stronger preference to the states $\mathbf{s}^{\tau}_t$ that contribute more to the learning strategy of the agent than the rest of the states. This means that our agent learns the policy $\pi_{\theta}$ based on the most important states $\mathbf{s}^{\tau}_t$. 

\subsection{Multi-Task Learner Component}

According to our architecture in Section~\ref{sec:arc} the multi-task learner optimizes the joint loss function $\mathcal{L}_{policy}$ to compute the parameters $w$ and $\theta$ of the policy component of Equations 3 and 5. In addition, based on the joint loss function  $\mathcal{L}_{learner}$ we calculate the parameters $\zeta$ of the multi-task learner component, and update the parameters $\eta$ and $\omega$ of the task importance component of Equations 6 and 7.

The input of the multi-task learner component is the $l_b$ state-action transitions, of each task $\tau$, stored in the replay buffer, and the weight matrix $\mathbf{M}$ generated by the task importance component. The multi-task learner component calculates the state-action value $Q(\mathbf{s}^{\tau}_t,\mathbf{a}^{\tau}_t)$, which is an approximation of the expected cumulative rewards of the agent, given the state $\mathbf{s}^{\tau}_t$ and action $\mathbf{a}^{\tau}_t$. We compute the state-action value $Q(s^{\tau}_t,a^{\tau}_t)$, as follows:

\begin{equation}
    Q(\mathbf{s}^{\tau}_t,\mathbf{a}^{\tau}_t) = \phi_{\zeta} (\mathbf{s}^{\tau}_t,\mathbf{a}^{\tau}_t) = MLP(\mathbf{s}^{\tau}_t \oplus \mathbf{a}^{\tau}_t)
\end{equation}
where $\zeta$ are the trainable parameters of the MLP function $\phi(\cdot)$, and $\oplus$ denotes the concatenation of the state $\mathbf{s}^{\tau}_t$ and action $\mathbf{a}^{\tau}_t$ vectors. Intuitively, the value $Q(\mathbf{s}^{\tau}_t,\mathbf{a}^{\tau}_t)$ corresponds to the benefit of the agent in terms of the expected reward for each task $\tau$, when taking the action $\mathbf{a}^{\tau}_t$ given the state $\mathbf{s}^{\tau}_t$ and following the policy $\pi_{\theta}$. By computing the value $Q(\mathbf{s}^{\tau}_t,\mathbf{a}^{\tau}_t)$ based on Equation 8, we can optimize the joint loss function $\mathcal{L}_{policy}$ with respect to the parameters $w$ and $\theta$ as follows \cite{lillicrap2019continuous,pmlr-v32-silver14}:

\begin{equation}
\begin{array}{c}

    w \leftarrow w - \alpha \nabla_{w} \mathcal{L}_{policy}(\pi_{\theta}) \\

    \theta \leftarrow \theta - \alpha \nabla_{\theta} \mathcal{L}_{policy}(\pi_{\theta}) \\

    \min\limits_{w,\theta}\mathcal{L}_{policy} = -\frac{1}{|\mathcal{T}|l_b} \sum_{\tau \in \mathcal{T}} \sum_{k=0}^{l_b} log \pi_{\theta}(\mathbf{a}^{\tau}_k, \mathbf{s}^{\tau}_k) [r(\mathbf{s}^{\tau}_k, \mathbf{a}^{\tau}_k) - M_{\tau,k} Q(\mathbf{s}^{\tau}_k, \mathbf{a}^{\tau}_k)] 
\end{array}
\end{equation}
where $\alpha$ is the learning rate. The term $[r(\mathbf{s}^{\tau}_k, \mathbf{a}^{\tau}_k) - M_{\tau,k} Q(\mathbf{s}^{\tau}_k, \mathbf{a}^{\tau}_k)]$ corresponds to the benefit of taking the action $\mathbf{a}^{\tau}_k$ given the state $\mathbf{s}^{\tau}_k$. The expected value $Q(\mathbf{s}^{\tau}_k, \mathbf{a}^{\tau}_k)$ is weighted by $M_{\tau,k}$ so as to strengthen/weaken the contribution of the state $\mathbf{s}^{\tau}_k$ when learning the policy $\pi_{\theta}$, accordingly. 

The joint loss function $\mathcal{L}_{learner}$ is formulated as a minimization mean squared error function with respect to parameters $\eta$ $\omega$ and $\zeta$ as follows:

\begin{equation}
    \begin{array}{c}
        \eta \leftarrow \eta - \alpha \nabla_{\eta} \mathcal{L}_{learner}(\pi_{\theta}) \\
        \omega \leftarrow \omega - \alpha \nabla_{\omega} \mathcal{L}_{learner}(\pi_{\theta}) \\
        \zeta \leftarrow \zeta - \alpha \nabla_{\zeta} \mathcal{L}_{learner}(\pi_{\theta}) \\
        
        \min\limits_{\eta,\omega,\zeta}\mathcal{L}_{learner} = \frac{1}{|\mathcal{T}|l_b} \sum_{\tau \in \mathcal{T}} \sum_{k=0}^{l_b} \bigg( r(\mathbf{s}^{\tau}_k,\mathbf{a}^{\tau}_k) -  M_{\tau,k} Q(\mathbf{s}^{\tau}_k,\mathbf{a}^{\tau}_k)\bigg)^2
        
    \end{array}
\end{equation}

Overall, to train our model we consider that the agent interacts with the environment in an episodic manner \cite{sutton2018reinforcement}. This means that the agent interacts with the environment within a finite horizon of $T$ interactions/events. We train our model for multiple episodes and optimize the joint loss functions $\mathcal{L}_{policy}$ and $\mathcal{L}_{learner}$ in Equations 9 and 10 with respect to the parameters $w$, $\theta$, $\eta$ $\omega$ and $\zeta$ through backpropagation with the Adam optimizer \cite{Diederik2015adam}.

\section{Experiments} \label{sec:exp}

\subsection{Setup}

\noindent \textbf{- Enviroment}. In our experiments, we evaluate the performance of the proposed model to select the timestamp $h$ of each event that maximizes the viewer's engagement $u_t$ and adoption $v_t$.  For each dataset we order the events according to the timestamps, and consider the first $70\%$ of the events as training set $\mathcal{E}^{train}$, $10\%$ for validation $\mathcal{E}^{val}$ and $20\%$ for testing $\mathcal{E}^{test}$.The agent interacts with an emulated environment\footnote{Provided the high risk that might hinder when evaluating the learned policy $\pi_{\theta}$ directly to the enterprises, in our study we perform off-line A/B testing based on the events of each dataset \cite{Gilotte2018AB,zou2019reinforcement}.} which models the behavioural policy $\pi_{\beta}$ of the events of each dataset. Following~\cite{Chen2019topk,Gilotte2018AB,zou2019reinforcement}, to emulate the behavioural policy $\pi_{\beta}$ we train a multi-head neural network on each dataset, which takes as input a sequence of events and outputs the average engagement and adoption of the next event. During the agent's training, we initialize the reinforcement learning environment with the events of the training set $\mathcal{E}^{train}$. To initialize the state $\mathbf{s}^{\tau}_t$ of the agent, we randomly select an event $e_t \in \mathcal{E}^{train}$ of the training set. At each step $t = 1, \ldots, T$, the agent takes an action $\mathbf{a}^{\tau}_t$ for each task $\tau$. Then, the agent receives the average engagement $u_t$ and adoption $v_t$ generated by the behavioural policy $\pi_{\beta}$ as a reward of each task. To evaluate the learned policy $\pi_{\theta}$, we initialize the reinforcement learning environment with the events of the test set $\mathcal{E}^{test}$. Similar to the training strategy, the state $\mathbf{s}^{\tau}_t$ of the agent is initialized by randomly selecting an event $e_t \in \mathcal{E}^{test}$ from the test set. The agent takes an action $\mathbf{a}^{\tau}_t$ and receives the reward by the multi-head network which models the behaviour policy $\pi_{\beta}$ of the test set $\mathcal{E}^{test}$.

\noindent \textbf{- Evaluation Metrics}. We evaluate the performance of our proposed model in terms of the step-wise variant of Normalized Capping Importance Sampling (NCIS) for each task as follows: \cite{swaminathan2015ncis,zou2019reinforcement}:

\begin{equation}
\begin{array}{l}
NCIS = \sum_{t=1}^{T} \frac{\Bar{\rho} r(\mathbf{s}^{\tau}_t, \mathbf{a}^{\tau}_t)} {\sum_{k=1}^{T} \Bar{\rho}} \\ \\ 

\Bar{\rho} = \min \{ \delta, \prod_{t=1}^{T} \frac{\pi_{\theta}(\mathbf{a}^{\tau}_t| \mathbf{s}^{\tau}_t)} {\pi_{\beta}(\mathbf{a}^{\tau}_t| \mathbf{s}^{\tau}_t)} \}

\end{array}
\label{eq:ncis}
\end{equation}
where $\Bar{\rho}$ is the max capping of the importance ratio, and $\delta$ is a threshold to ensure small variance and control the bias of the policy $\pi_{\theta}$ towards the behavioural policy $\pi_{\beta}$. The term $\Bar{\rho}r(\mathbf{s}^{\tau}_t, \mathbf{a}^{\tau}_t)$ is the capped importance weighted reward of a task $\tau$. Intuitively, by adopting different rewards in the term $\Bar{\rho}r(\mathbf{s}^{\tau}_t, \mathbf{a}^{\tau}_t)$, we can measure the performance of the policy $\pi_{\theta}$ to approximate the behavioural policy $\pi_{\beta}$. By setting each reward $r(\mathbf{s}^{\tau}_t, \mathbf{a}^{\tau}_t)$ equal to the viewer's engagement and adoption as in Section 3, we can evaluate the performance of the proposed model based on the respective metrics Eng. NCIS and Ad. NCIS for both tasks. As the emulated environment is initialized randomly, we repeated our experiments five times and report average Eng. NCIS and Ad. NCIS in our experiments. 

\noindent \textbf{- Baselines}. We compare the proposed MERLIN model against the following strategies: FeedRec~\cite{zou2019reinforcement}, AMT\footnote{\url{https://github.com/braemt/attentive-multi-task-deep-reinforcement-learning}} \cite{Timo2020AMT}, IMPALA\footnote{\url{https://github.com/deepmind/scalable\_agent}} \cite{lasse2018impala} and PopART \cite{Hessel2019Popart}. As there are no publicly available implementations of FeedRec and PopART, we implemented both from scratch and published our source codes\footnote{\url{https://github.com/stefanosantaris/merlin}}.

\noindent \textbf{- Parameter Configuration}. For each examined model, we tuned the hyperparameters on the validation set, following a grid-selection strategy. In FeedRec, we set the state representation dimensionality $d_s=256$ for Enterprises 1 and 3, and $d_s=128$ for Enterprises 2 and 4. At the $t$-th step, the FeedRec model takes as an input all the events occurred prior to the current step with $l=0$. In AMT we fix a $d_s=128$ dimensional state representation for all datasets, with a time window $l=30$ previous events. In IMPALA and PopART the state representation's dimensionality is fixed to $d_s=64$ for all Enterprises. The window length $l$ in IMPALA and PopART is set to $20$ and $23$, respectively. In the proposed MERLIN model we use a $d_s=128$ dimensional state representation for Enterprises 1 and 4, and $256$ and $64$ for Enterprises 2 and 3, respectively. The window length $l$ is fixed to $10$ for  Enterprise 1, and $15$ for Enterprises 2, 3 and 4. In addition, the size of the replay buffer $l_b$ is set to 128 for all Enterprises. In all the examined models, we follow an $\epsilon$-greedy exploration-exploitation strategy and set $\epsilon=0.1$. The discount factor $\gamma$ is fixed to $0.92$ and the learning rate is set to $\alpha=0.001$. In the emulated environment, we set the number of interactions/events to $200$ and the number of episodes to $300$. 

All our experiments were conducted on a single server with an Intel Xeon Bronze 3106, 1.70GHz CPU. The operating system of the server was Ubuntu 18.04 LTS. We accelerated the training of the model using the GPU Geforce RTX 2080 Ti graph card. Our proposed MERLIN model was implemented in Pytorch 1.7.1 and we created the reinforcement learning environment with the OpenAI Gym 0.17.3 library. 

\subsection{Performance Evaluation}

\begin{table*}
    \caption{Performance comparison of the examined models on the engagement and adoption tasks in terms of average Eng.~NCIS and Ad. NCIS. Bold values indicate the best method using a statistical significance t-test with $p<0.01$.}
    \centering
    \begin{tabular}{c|c|cccc}
        \hline
        \textbf{Task} & \textbf{Model} & \multicolumn{4}{c}{\textbf{Datasets}} \\ \hline
        & & \textbf{E1} & \textbf{E2} & \textbf{E3} & \textbf{E4} \\ \hline
        \multirow{5}{*}{\textbf{Avg. Eng. NCIS}}    & \textbf{FeedRec}  & $0.553$ & $0.591$ & $0.423$ & $0.467$ \\
                                                & \textbf{AMT}      & $0.462$ & $0.513$ & $0.371$ & $0.380$ \\
                                                & \textbf{IMPALA}   & $0.452$ & $0.493$ & $0.352$ & $0.314$ \\
                                                & \textbf{PopART}   & $0.421$ & $0.460$ & $0.432$ & $0.401$ \\
                                                & \textbf{MERLIN}   & \textbf{0.622} & \textbf{0.663} & \textbf{0.512} & \textbf{0.552} \\ \hline \hline
        \multirow{5}{*}{\textbf{Avg. Ad. NCIS}}      & \textbf{FeedRec}  & $8.122$ & $15.271$ & $14.393$ & $27.292$ \\
                                                & \textbf{AMT}      & $6.284$ & $12.781$ & $11.842$ & $20.962$ \\
                                                & \textbf{IMPALA}   & $5.023$ & $10.523$ & $9.232$ & $18.284$ \\
                                                & \textbf{PopART}   & $4.891$ & $9.362$ & $9.013$ & $16.642$ \\
                                                & \textbf{MERLIN}   & \textbf{10.112} & \textbf{17.292} & \textbf{16.961} & \textbf{29.554} \\
        
    \end{tabular}
    \label{tab:ncis}
\end{table*}

In Table \ref{tab:ncis}, we evaluate the performance of the examined models in terms of   average Eng.~NCIS and Ad. NCIS  over the five trials in the emulated environment for the engagement and adoption tasks, respectively. The proposed MERLIN model significantly outperforms the baselines in all datasets. This indicates that MERLIN can efficiently learn a common policy $\pi_{\theta}$ that optimizes both tasks concurrently. Compared with the second best method FeedRec, MERLIN achieves relative improvements of $15.76$ and $15.96\%$ in terms of Eng.~NCIS and Ad. NCIS, respectively. FeedRec performs better than the other baseline approaches because FeedRec formulates a joint loss function for training the agent on the different tasks. However, each task in FeedRec contributes equally when learning the policy $\pi_{\theta}$, and therefore the agent ignores the evolutionary patterns and the importance of the state-action transitions for each task. The proposed MERLIN model overcomes this problem by integrating the training parameters of the task importance component in the common learning strategy of the policy and multi-task learner components. In doing so, MERLIN balances the contribution of each task to the generated policy. 

\begin{figure*}[t]
    \centering
    \includegraphics[scale=0.118]{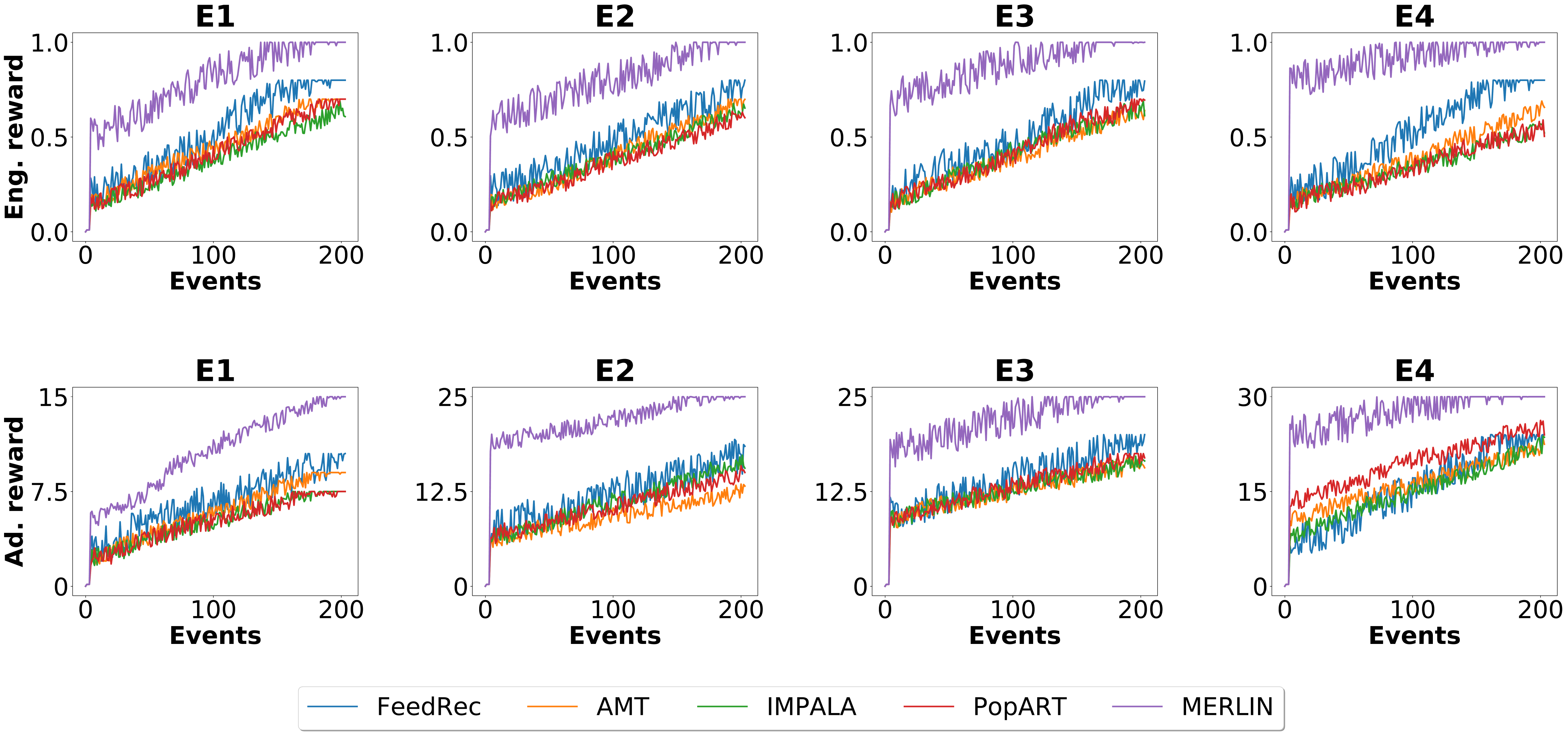} 
    
    \caption{The Eng. reward and Ad.reward based on Equations \ref{eq:engagement} and \ref{eq:adoption} of the examined models for the engagement and adoption tasks, when the interactions/events evolve in the emulated environment.}
    \label{fig:rewards}
\end{figure*}

In Figure \ref{fig:rewards} we report the Eng. reward and Ad. reward based on Equations \ref{eq:engagement} and \ref{eq:adoption} for the engagement and adoption tasks, respectively, when the interactions/events evolve in the emulated environment. We observe that MERLIN constantly achieves higher rewards than the other baseline approaches at the first interactions. This demonstrates the effectiveness of MERLIN to weigh the importance of each task during training and learn a policy that optimizes both tasks. In addition, we observe that the Ad.~reward in the adoption task of MERLIN converges faster in Enterprises 2, 3 and 4 than in Enterprise 1. As discussed in Section \ref{sec:data}, the viewer's adoption in Enterprises 2, 3 and 4 increase over time. Therefore, the task importance component promotes the adoption task during the training of the policy, thus achieving high reward in Enterprises 2, 3 and 4 at the beginning of the interactions.



\subsection{Multi-Task vs Single-Task Learning in Parameter Configuration}
\begin{figure*}[t]
    \centering
    \includegraphics[scale=0.118]{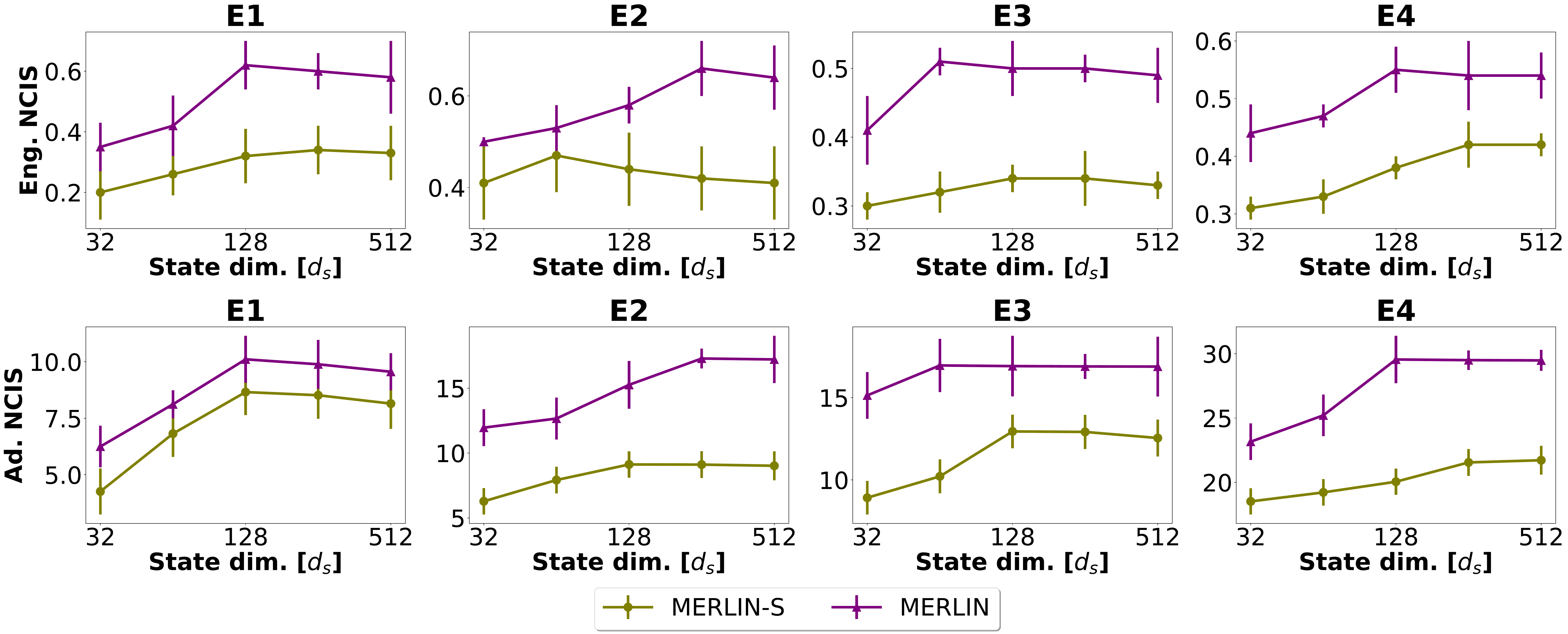}
    \caption{Impact of the state representation's dimensionality $d_s$ on the performance of MERLIN and its single-task variant MERLIN-S for the engagement and adoption tasks.}
    \label{fig:engagement_adoption_embeddings}
\end{figure*}

\begin{figure*}[t]
    \centering
    \includegraphics[scale=0.118]{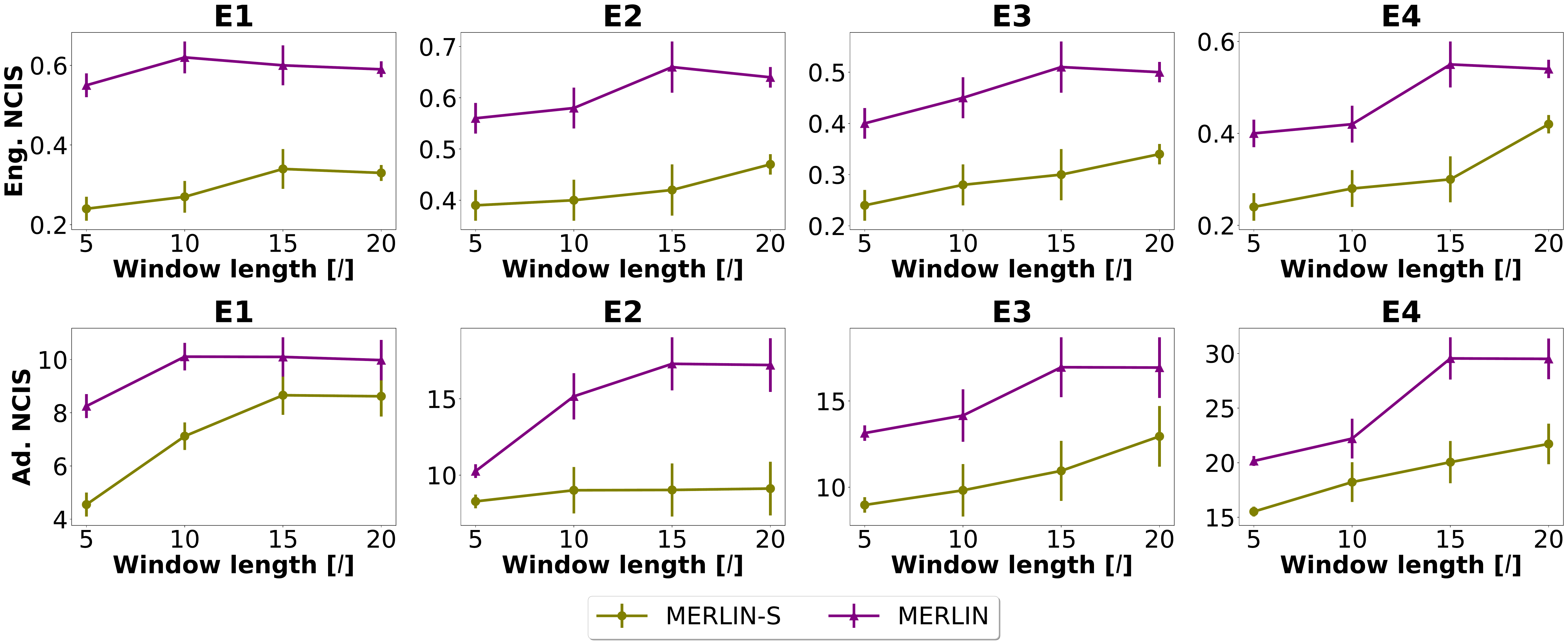}
    \caption{Impact of the window length $l$ on MERLIN and MERLIN-S.}
    \label{fig:engagement_adoption_window}
\end{figure*}

In the next set of experiments we compare the proposed MERLIN model with its variant MERLIN-S. In particular, the agent of the variant MERLIN-S is trained on a single task, ignoring the multi-task learning strategy of MERLIN. In Figure~\ref{fig:engagement_adoption_embeddings}, we study the impact of the state representation's dimensionality $d_s$ on the performances of MERLIN and MERLIN-S in terms of Eng.~NCIS and Ad.~NCIS for the engagement and adoption tasks, when varying $d_s$ in $\{32, 64, 128, 256, 512\}$.  We observe that MERLIN achieves the best performance when setting $128$ dimensions for Enterprises 1 and 4, $256$ for Enterprise 2, and $64$ for Enterprise 3. By increasing the dimensionality $d_s$ of the state representation, the agent of MERLIN achieves similar performances in both tasks. We observe that MERLIN significantly outperforms the MERLIN-S model in both tasks, indicating the importance of the multi-task learning strategy to efficiently extract knowledge from both tasks. In Figure \ref{fig:engagement_adoption_window}, we present the impact of the window length $l$ on MERLIN and MERLIN-S. We vary the window length $l$ from $5$ to $20$ by a step of $5$. MERLIN requires $10$ past events in Enterprise 1, and $15$ events in Enterprises 2, 3 and 4. Moreover, we observe that MERLIN constantly outperforms the single task variant MERLIN-S. Notice that MERLIN-S achieves the best performance when the window length $l$ is set to $15$ past events for  Enterprise 1, and $20$ for Enterprises 2, 3 and 4. Therefore, MERLIN-S requires a higher window length $l$ than   MERLIN  in all Enterprises, as MERLIN-S omits the auxiliary information of the other task when training the agent.



\section{Conclusions} \label{sec:conclusion}

In this study, we presented a multi-task reinforcement learning strategy to train an agent so as to select the optimal time of a live video streaming event in large enterprises, aiming to improve the viewer's engagement and adoption. In the proposed MERLIN model, we formulate the engagement and adoption tasks as different MDPs and design a joint loss function to extract knowledge from both tasks. To determine the contribution of each task to the training strategy of the agent, we implement a task importance learner component that extracts the most important information, that is the most important state-action transitions from the replay buffer based on the Transformer's architecture. Having weighted the transitions, the agent of MERLIN learns a common policy for both tasks. Our experiments with four real-world datasets demonstrate the superiority of our model against several baseline approaches in terms of viewer's engagement and adoption. The proposed MERLIN model can significantly help enterprises in selecting the optimal time of an event. Provided that nowadays the majority of the events are online, the enterprises want to ensure that their employees/viewers adopt the video streaming events with high engagement. This means that with the help of MERLIN in scheduling the live video streaming events, the enterprises can communicate with their employees efficiently, which as a consequence reflects on significant productivity gains \cite{ibmadoptionreport}. An interesting future direction is to study the influence of  distillation strategies on the proposed MERLIN model~\cite{parisottoefficient}. 

%
%
%
%
%
\bibliographystyle{splncs04}
\bibliography{mybibliography}
%




\end{document}